%% file: main_for_iccv.tex
\documentclass[10pt,twocolumn,letterpaper]{article}

\usepackage{iccv}
\usepackage{times}
\usepackage[table]{xcolor} 
\usepackage{graphicx}
\usepackage{amsmath}
\usepackage{amssymb}
\usepackage{booktabs}
\usepackage{subfiles}
\usepackage{amsmath,amssymb,tabularx}
\usepackage{multirow}
\usepackage{cite}
\usepackage{xcolor,pifont}
\usepackage{fontawesome}
\usepackage{arydshln}
\usepackage[english]{babel}
\usepackage{csquotes}{\itshape}
\usepackage[accsupp]{axessibility} 

\definecolor{LightCyan}{rgb}{0.88,1,1}
\definecolor{rowred}{rgb}{0.35,1,1}

\definecolor{rowlightblue}{rgb}{1,0.88,1}
\definecolor{rowblue}{rgb}{1,0.55,1}

\newcommand{\blue}{\cellcolor{blue!25}}
\newcommand{\gray}{\cellcolor{gray!25}}
\newcommand{\red}{\cellcolor{red!25}}
\NewDocumentCommand{\rot}{O{45} O{1em} m}{\makebox[#2][l]{\rotatebox{#1}{#3}}}%

\usepackage[pagebackref=true,breaklinks=true,letterpaper=true,colorlinks,bookmarks=false]{hyperref}

\newcommand{\Paragraph}[1]{\vspace{1mm} \noindent \textbf{#1.} \hspace{0mm}}
\newcommand{\Section}[1]{\vspace{-1mm} \section{#1} \vspace{-1mm}}

\usepackage[capitalize]{cleveref}
\crefname{section}{Sec.}{Secs.}
\Crefname{section}{Section}{Sections}
\Crefname{table}{Table}{Tables}
\crefname{table}{Tab.}{Tabs.}
\DeclareMathOperator*{\argmin}{arg\,min}

\iccvfinalcopy 

\def\iccvPaperID{5233} 

\begin{document}

\title{Learning Clothing and Pose Invariant 3D Shape Representation \\for Long-Term Person Re-Identification}

\author{Feng Liu, Minchul Kim, ZiAng Gu, Anil Jain, Xiaoming Liu \\
Department of Computer Science and Engineering \\
Michigan State University, East Lansing MI 48824\\
{\tt \{liufeng6,kimminc2,guziang,jain,liuxm\}@msu.edu}
}

\maketitle


\subfile{sec_0_abstract.tex}


\subfile{sec_1_intro.tex}
%
\subfile{sec_2_priors.tex}

%
\subfile{sec_3_method.tex}

\subfile{sec_4_exp.tex}

%

\subfile{sec_5_conclusion.tex}

{\small
\bibliographystyle{ieee_fullname}
\bibliography{egbib}
}

\end{document}

%% file: sec_0_abstract.tex
\begin{abstract}

Long-Term Person Re-Identification (LT-ReID) has become increasingly crucial in computer vision and biometrics. In this work, we aim to extend LT-ReID beyond pedestrian recognition to include a wider range of real-world human activities while still accounting for cloth-changing scenarios over large time gaps. This setting poses additional challenges due to the geometric misalignment and appearance ambiguity caused by the diversity of human pose and clothing. To address these challenges, we propose a new approach \textbf{3DInvarReID} for (i) disentangling identity from non-identity components (pose, clothing shape, and texture) of 3D clothed humans, and (ii) reconstructing accurate 3D clothed body shapes and learning discriminative features of naked body shapes for person ReID in a joint manner. To better evaluate our study of LT-ReID, we collect a real-world dataset called CCDA, which contains a wide variety of human activities and clothing changes. Experimentally, we show the superior performance of our approach for person ReID. Code is available at \url{http://cvlab.cse.msu.edu/project-reid3dinvar.html}.


\end{abstract}


%% file: sec_1_intro.tex
\section{Introduction}\label{sec:intro}

Person Re-Identification (ReID) aims to recognize and match a specific pedestrian in various locations and at different times~\cite{liu2023farsight,ye2021deep,ahmed2015improved,zheng2015scalable,li2018harmonious}.
This is a crucial task for various applications, including crime prevention, forensic identification and security monitoring~\cite{gong2011person,zheng2016person}. 

\begin{figure}[t]
  \centering
   \includegraphics[width=0.98\linewidth]{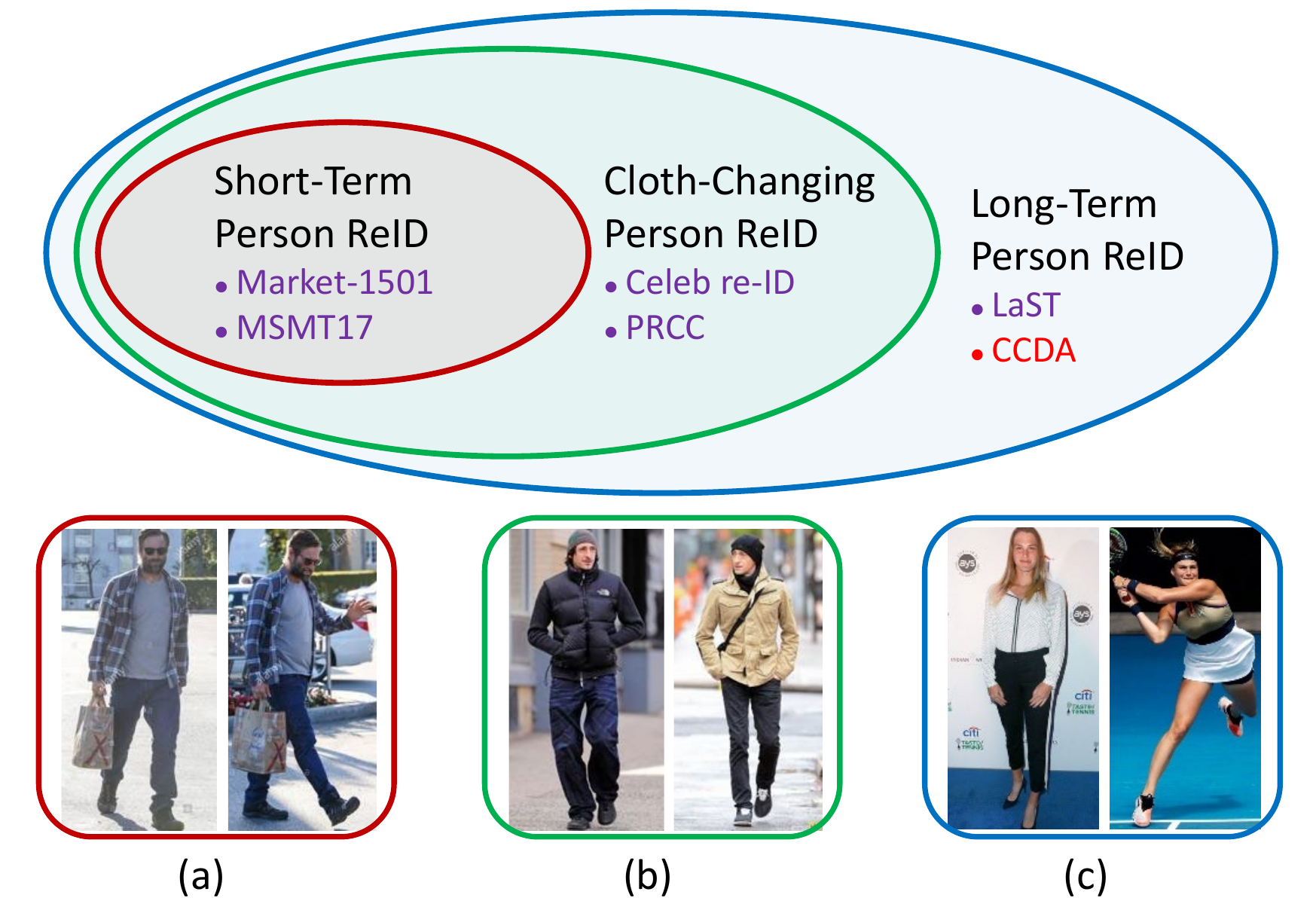}
   \vspace{0mm}
   \caption{\small Illustration of the differences between various person re-identification (ReID) settings. Both (a) 
   \emph{conventional/short-term} and (b) \emph{cloth-changing} person ReID benchmarks often restrict subjects to walking or standing, limiting their applications in real-world scenarios. This paper expands on the long-term person re-identification (LT-ReID) setting by tackling a wider range of human activities, increasing its practicality.}
   \label{fig:examples}
\end{figure}

Most existing works~\cite{ge2020mutual,ge2020self,li2018unsupervised,li2019unsupervised} in this field concentrate on the short-term scenarios, assuming that pedestrians' clothing remains unchanged.
However, in this paper, we focus on a more challenging yet practical scenario of Long-Term Person Re-Identification (LT-ReID), where the objective is to recognize individuals over long time periods while taking into account \textbf{variations in clothing} and \textbf{diverse human activities}.
For the first time, we extend person re-identification beyond pedestrian recognition to encompass a wider range of human activities, such as identifying students playing tennis or soldiers crawling in the field (see Fig.~\ref{fig:examples}).
%
\emph{This setting poses new challenges due to the geometric misalignment and appearance ambiguity caused by the diversity of human poses and their clothing.}

Recently, various approaches~\cite{gu2022clothes,huang2019beyond,yu2020cocas,huang2019celebrities,yang2019person,shu2021large,li2021learning,jin2022cloth} have been proposed to investigate LT-ReID under clothing changes. They extract clothing-irrelevant features for robust person ReID by custom-designed architectures~\cite{huang2019beyond,huang2021clothing}, training process~\cite{li2021learning}, loss functions~\cite{gu2022clothes}, and data augmentation~\cite{zheng2019joint}.
However, these methods only attempt to mine texture-insensitive body-structural cues in $2$D space while ignoring the prior knowledge that the human body is a $3$D non-rigid object.
A new line of research introduces $3$D priors for LT-ReID by either lifting $2$D images to a $3$D space~\cite{zheng2020parameter} or including $3$D body reconstruction as an auxiliary task~\cite{chen2021learning}.
However, without modeling the $3$D clothing, the clothing-sensitive features can not be properly disentangled in either method.
Moreover, none of the methods above handle body images with diverse activities.

Given the numerous variations in body images, including body pose, clothing, and view angles, we posit that the most reliable identity cue for LT-ReID is the $3$D naked (unclothed) body shape, if it can be accurately and discriminately estimated from a $2$D body image.
%
Obviously  this is extremely challenging due to confounding factors and the lack of supervision, such as paired images and $3$D naked body scans. However, taking inspiration from advancements in $3$D feature learning for face recognition~\cite{peng2017reconstruction,liu2018disentangling}, we propose a new algorithm, \textbf{3DInvarReID}, to disentangle identity (naked body) from non-identity components (pose, clothing shape and texture) of $3$D clothed humans. 
This innovative approach not only reconstructs accurate $3$D clothed body shapes that faithfully represent the input $2$D images, but it also simultaneously learns discriminative naked shape features that effectively enhance LT-ReID.

An effective representation of the $3$D shape and texture of the human body is a key component of such a learning-based process. 
To this end, we propose a joint two-layer neural implicit function to represent 3D humans, where identity, clothing shape, and texture components are disentangled into latent representations. 
Based on the composite model, we jointly learn a model fitting module to disentangle identity from non-identity components (body pose, clothing shape and texture) from $2$D images. 
Modeling texture, along with a differentiable renderer enables us to compare the rendered image with the input image in a self-supervised manner. This allows the learning process to be supervised by both image reconstruction loss and identification loss, using a set of $2$D images with identity labels {\it only}.
Comprehensive experiments demonstrate the superiority of our method in diverse ReID benchmarks.
%
Additionally, to advance the research in the field of LT-ReID, we collect a Cloth-Changing and Diverse Activities (CCDA) dataset (see Fig.~\ref{fig:examples}). The CCDA dataset is specifically designed to evaluate the ReID of the person undergoing both human activities and changes in clothing.

In summary, the contributions of this work include:

$\diamond$ 
We propose a novel LT-ReID method, \textbf{3DInvarReID}, to  learn clothing/pose invariant $3$D shape representation.

$\diamond$
We devise a novel joint two-layer implicit model that fully models a textured $3$D clothed human. Our approach includes a robust and discriminative fitting process that disentangles identity and non-identity features in reconstructing two-layer $3$D body shapes from real-world images. 


$\diamond$
We achieve superior performance in both LT-ReID accuracy and $3$D body shape reconstruction.


%% file: sec_2_priors.tex

\begin{table}[t!]
\newcommand{\tabincell}[2]{\begin{tabular}{@{}#1@{}}#2\end{tabular}}
\newcommand{\greencheck}{\textcolor{green}{\ding{51}}} 
\newcommand{\redX}{\textcolor{red}{\ding{55}}}
\newcommand{\halfcheck}{\textcolor{green}{\ding{51}}{\small\textcolor{black}{\msquare }}}
\begin{center}
\resizebox{1\linewidth}{!}{
\begin{tabular}{l |c| c | c | c  }
\hline
Method & \tabincell{c}{End-to-End \\ trainable} & \tabincell{c}{Model\\ texture}   & Discriminative & \tabincell{c}{Model\\ type}    \\

\hline\hline
\multicolumn{5}{c}{\emph{$3$D modeling methods}} \\ 
\hdashline
SCANimate~\cite{saito2021scanimate}  & - & \redX  & \redX & universal  \\
SMPLicit~\cite{corona2021smplicit}   & - & \redX  & \redX & universal  \\
Neural-GIF~\cite{tiwari2021neural}   & - & \redX  & \redX & individual \\
SNARF~\cite{chen2021snarf} & - & \redX  & \redX  & individual\\
gDNA~\cite{chen2022gdna} & - & \redX  & \redX  & universal  \\

\hline
\multicolumn{5}{c}{\emph{$3$D fitting methods}} \\ 
\hdashline
PiFu~\cite{saito2019pifu}  & \redX & \greencheck  & \redX & -  \\
PiFuHD~\cite{saito2020pifuhd}  & \redX & \redX  & \redX & - \\
Arch~\cite{huang2020arch}  & \redX & \greencheck  & \redX & - \\
Arch++~\cite{he2021arch++}  & \redX & \greencheck  & \redX & - \\
ICON~\cite{xiu2022icon}  & \redX & \greencheck  & \redX & - \\
ClothWild~\cite{moon20223d} & \redX & \redX  & \redX  & - \\
PHORHUM~\cite{alldieck2022photorealistic}  & \greencheck & \greencheck  & \redX  & - \\
\hline
\textbf{3DInvarReID}  & \greencheck & \greencheck  & \greencheck  & universal \\
\hline
\end{tabular}
}
\end{center}
\vspace{-1mm}
\caption{\small Overview of the $3$D clothed human modeling (top) and fitting (bottom) methods. Our method is the only one that models clothing texture and learns discriminative information compared to 3D modeling methods. Compared to $3$D fitting methods, our end-to-end trainable pipeline enables disentangling identity-sensitive shape features from whole-body images.}
\label{tab:3D_modeling_review}
\end{table}

\begin{figure*}[t]
  \centering
  \includegraphics[width=0.92\linewidth]{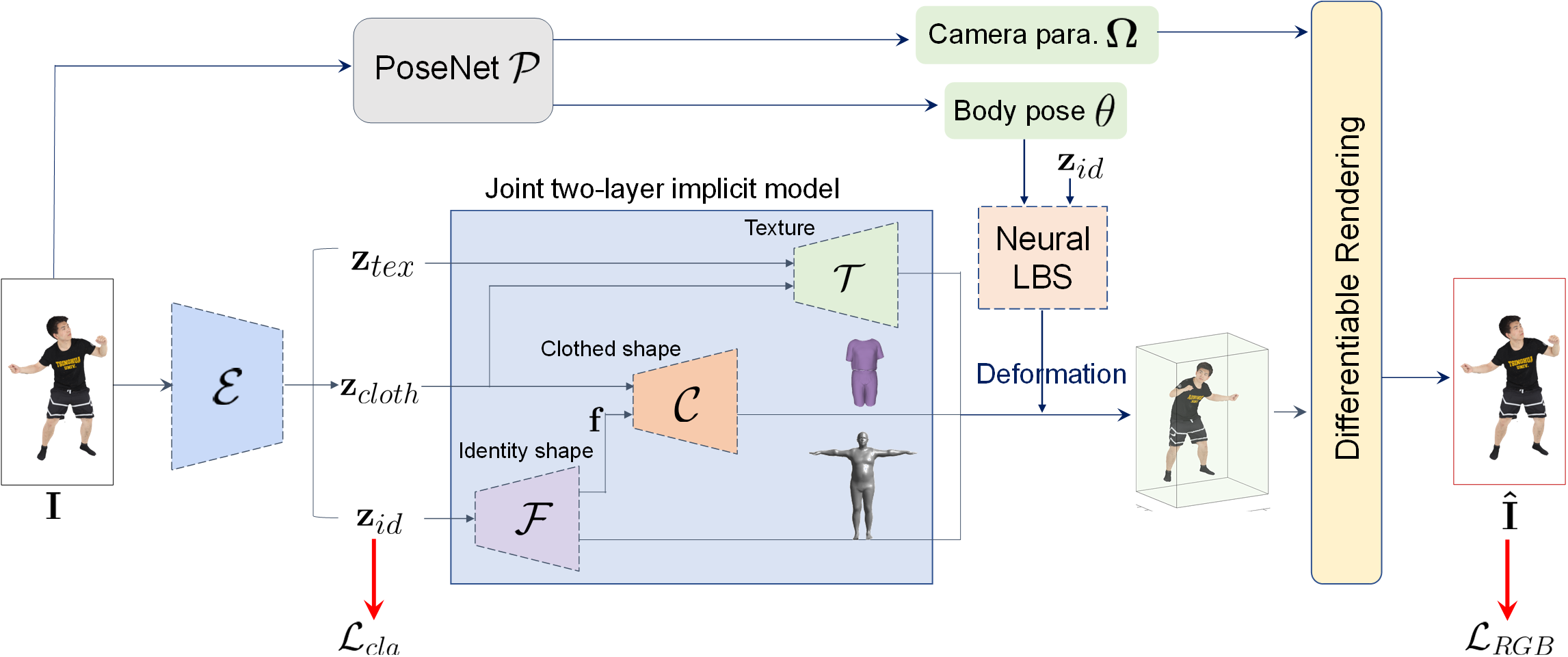}
   \caption{\small Overview of the proposed joint learning framework for long-term person re-identification and $3$D clothed body shape reconstruction. During the inference of ReID, the identity shape feature $\mathbf{z}_{id}$ is utilized for matching.}
   \label{fig:flowchart}
\end{figure*}

\section{Prior Work}\label{sec:prior}

\Paragraph{Person Re-identification}
Person ReID aims to match a person across images captured by a distributed camera system. 
The majority of prior methods~\cite{ge2020mutual,ge2020self,li2018unsupervised,li2019unsupervised,lin2019bottom,wang2018transferable,yu2019unsupervised,zhai2020ad} assume a short-term application scenario without clothing changes by the person. 
%
This limitation has generated a growing interest in long-term cloth-changing person ReID~\cite{gu2022clothes,yang2019person,li2021learning,jin2022cloth}. 
Datasets such as Real28~\cite{wan2020person}, VC-Clothes~\cite{wan2020person}, PRCC~\cite{yang2019person}, LTCC~\cite{shu2021large}, COCAS~\cite{yu2020cocas} and Celebrities-reID~\cite{huang2019celebrities,huang2019beyond} are collected to facilitate this research. 
%
These datasets, however, either ignore or only minimally consider human activities, assuming that subjects are pedestrians with a restricted set of activities, limiting their applicability in real-world scenarios. As a result, there is a noticeable discrepancy between published approaches and the real-world LT-ReID problem.
In contrast to the focus on the clothing-change person ReID, our research takes a step further by addressing a more challenging and practical issue of person ReID that involves diverse human activities, which are not limited to walking.


\Paragraph{3D Clothed Human Modeling and Fitting}
In early attempts~\cite{jiang2020bcnet,ma2020learning,patel2020tailornet}, a clothed person was modeled as displacements over naked body meshes, obtained by SMPL~\cite{loper2015smpl}. 
However, the fixed mesh topology and bounded resolution approach limit geometric expressivity.
Recently, neural implicit representations have been explored to model $3$D body shapes due to their topological flexibility and resolution independence~\cite{chen2021snarf,palafox2021npms,saito2021scanimate,corona2021smplicit,tiwari2021neural,wang2021metaavatar,chen2022gdna,shao2022doublefield}.
However, as shown in Tab.~\ref{tab:3D_modeling_review}, modeling texture in $3$D remains a challenge. 
While these approaches provide rich geometric detail, insufficient attention has been paid to the discriminativeness of the resulting  body shapes. 
In contrast, we build a \emph{discriminative} and \emph{textured} $3$D clothed human, serving the purpose of LT-ReID.

These $3$D clothed models can be naturally applied to monocular $3$D reconstruction ($2$D-to-$3$D fitting). 
Generally, the input image is encoded as a latent vector, from which the generative model reconstructs the $3$D shape~\cite{liu20222d,liu2021fully,corona2021smplicit,moon20223d}.
Alternatively, two-step pipelines~\cite{saito2019pifu,saito2020pifuhd,huang2020arch,he2021arch++,xiu2022icon} firstly recover $2.5$D
sketches (\emph{e.g.}, surface normal), and then infer a full $3$D shape. 
A common limitation of these works is that they require $3$D body scans for training, as they are trained on synthetic datasets derived from $3$D scans and their rendered images.
Furthermore, existing methods do not explicitly consider the discriminative ability of reconstructed $3$D clothed body shapes. 
LVD~\cite{corona2022learned} and SHAPY~\cite{choutas2022accurate} introduce new discriminative fitting pipelines to reconstruct naked body shapes from images. 
However, without modeling $3$D clothing, $2$D image cues can not be fully exploited.     

We propose a novel joint two-layer shape and texture representation of a $3$D clothed human model, consisting of both shape and texture.
Together with a model fitting module, our representation allows semi-supervised training from images without $3$D labels. 
More importantly, guided by the completed $3$D model and the discriminative $2$D-to-$3$D fitting module, our approach disentangles identity-features from identity-irrelevant features in $3$D space for LT-ReID.
Tab.~\ref{tab:3D_modeling_review} compares our method with prior works.


%% file: sec_3_method.tex

\section{Proposed Method}\label{sec:method}

\subsection{Problem Formulation}
A $3$D clothed human model is described by three disentangled latent variables: identity shape, clothing shape and clothing texture. 
As shown in Fig.~\ref{fig:flowchart}, these latent representations can be sequentially decoded into \emph{canonical} $3$D shape and texture, respectively by three decoders.
To enable self-supervised training on real images, we estimate these latent codes along with the body pose and camera projection parameters. 
In this work, we use an off-the-shelf method~\cite{moon2022accurate} as our PoseNet to predict pose and camera projection, while our image encoder focuses on identity disentanglement learning, \emph{i.e.}, the fitting module.  
These networks disentangle identity and non-identity components of $3$D shapes and reconstruct the input body images via a differentiable render.

Formally, given a training set of $T$ images $\{\mathbf{I}_i\}_{i=1}^{T}$ and the corresponding identity labels $\{l_i\}_{i=1}^{T}$,
the image encoder $\mathcal{E}(\mathbf{I}):\mathbf{I}\xrightarrow{} \mathbf{z}_{id}, \mathbf{z}_{cloth}, \mathbf{z}_{tex}$ predicts the identity shape code of naked body $\mathbf{z}_{id}\in\mathbb{R}^{L_{id}}$, clothed shape code $\mathbf{z}_{cloth}\in\mathbb{R}^{L_{cloth}}$ and texture code $\mathbf{z}_{tex}\in\mathbb{R}^{L_{tex}}$. Functions $\mathcal{F}$, $\mathcal{C}$ and $\mathcal{T}$ decode the latent codes to identity shape, clothing shape and texture components, respectively. 
Additionally, PoseNet $\mathcal{P}$ predicts the camera projection parameters $\mathbf{\Omega}$ and SMPL body pose $\mathbf{\theta}$: $(\mathbf{\Omega}, \mathbf{\theta}) = \mathcal{P}(\mathbf{I})$.

Mathematically, the learning objective is defined as:
\begin{equation}
\label{eqn:objective}
\argmin_{\mathcal{E},\mathcal{F},\mathcal{C},\mathcal{T}} \sum_{i=1}^T \left(\left|\hat{\mathbf{I}}_i -  \mathbf{I}_i\right|_1 
+ \mathcal{L}_{cla}(\mathbf{z}_{id},l_i)\right),
\end{equation}
where $\mathcal{L}_{cla}$ is the classification loss. $\hat{\mathbf{I}}$ is the rendered image.
This  objective enables us to jointly learn accurate $3$D clothed shape and discriminative shape for the naked body.

\subsection{Joint Two-Layer Implicit Model}
We jointly model $3$D naked body shape, clothed shape and texture in a \emph{canonical} space by implicit representations.

\Paragraph{Discriminative Body Shape Component}
We represent the $3$D naked body shape as the $\tau=0.5$ level set of the occupancy function~\cite{mescheder2019occupancy}:
\begin{equation}
\label{eqn:naked}
\mathbf{S}_{id}(\mathbf{z}_{id}) = \{\mathbf{x} |\mathcal{F}(\mathbf{z}_{id}, \mathbf{x})=\tau\},
\end{equation}
where $\mathcal{F}$ predicts the occupancy value, $o_1$ for any point $\mathbf{x}$ in the canonical space. Specifically,
\begin{equation}
\label{eqn:naked_func}
\mathcal{F}: \mathbb{R}^{L_{id}} \times \mathbb{R}^{3} \xrightarrow{} (o_1, \mathbf{f}),
\end{equation}
where $\mathbf{f}\in\mathbb{R}^{L_{f}}$ is a point-wise feature and will be utilized to predict clothing details.

\begin{figure}[t]
  \centering
  \includegraphics[width=0.98\linewidth]{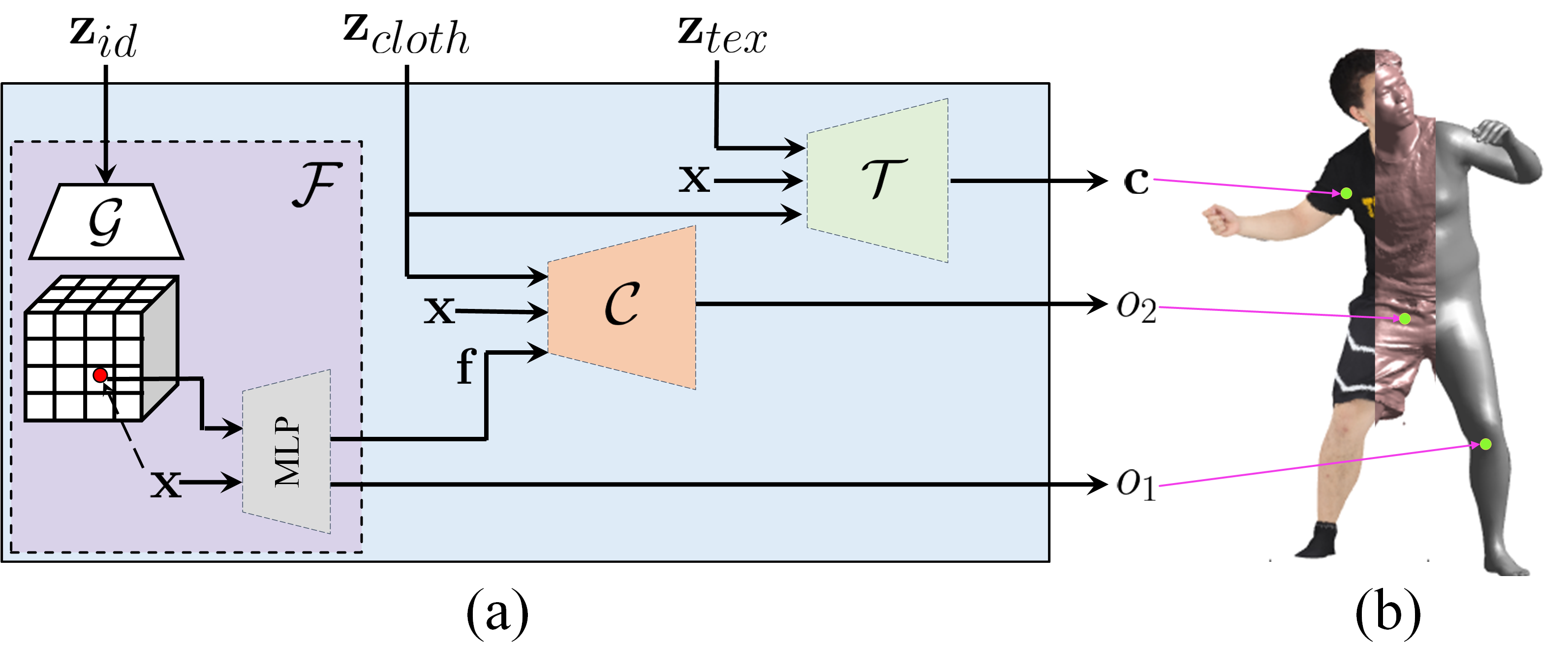}
   \caption{\small Joint two-layer implicit model. The naked body shape model $\mathcal{F}$ and clothed body shape model $\mathcal{C}$ take identity shape code $\mathbf{z}_{id}$, clothing shape code $\mathbf{z}_{cloth}$ and a spatial point $\mathbf{x}$, and produces two occupancy values $o_1$ and $o_2$. The texture model $\mathcal{T}$ takes $\mathbf{z}_{tex}$ and $\mathbf{z}_{cloth}$ to estimate RGB color at $\mathbf{x}$. A $3$D generator $\mathcal{G}$ uses $\mathbf{z}_{id}$ to produce a $3$D feature volume, enabling hierarchical point-wise feature representation.} 
   \label{fig:implicit}
\end{figure}

Following~\cite{chen2022gdna}, we make use of function $\mathcal{F}$, implemented via a Multi-Layer Perceptron (MLP), coupled with a $3$D CNN-based generator $\mathcal{G}$ to model $3$D naked bodies.
%
As shown in Fig.~\ref{fig:implicit}\textcolor{red}{(a)}, the generator produces a $3$D feature volume using $\mathbf{z}_{id}$ as input. We then use trilinear interpolation to query continuous $3$D points and feed the feature at $\mathbf{x}$ to the MLP.  

\Paragraph{Clothed Shape Component}
Similarly, we also represent the clothed body shape as the $\tau=0.5$ level set function:
\begin{equation}
\label{eqn:clothed}
\mathbf{S}_{cloth}(\mathbf{z}_{cloth}) = \{\mathbf{x} |\mathcal{C}(\mathbf{z}_{cloth}, \mathbf{f}, \mathbf{x})=\tau\},
\end{equation}
where $\mathcal{C}$ is implemented as a MLP:
\begin{equation}
\label{eqn:clothed_func}
\mathcal{C}: \mathbb{R}^{L_{cloth}}\times\mathbb{R}^{L_{f}} \times\mathbb{R}^{3} \xrightarrow{} o_2.
\end{equation}
$\mathcal{C}$ outputs the occupancy value $o_{2}$ to represent the clothed shape information.


\Paragraph{Texture Component}
We define a texture field as a mapping function $\mathcal{T}$ from a point $\mathbf{x}$ in the canonical space, texture latent $\mathbf{z}_{tex}$ and $\mathbf{z}_{cloth}$ to a RGB value $\mathbf{c}\in\mathbb{R}^3$:
\begin{equation}
\label{eqn:texture_func}
\mathcal{T}: \mathbb{R}^{L_{tex}}\times\mathbb{R}^{L_{cloth}} \times\mathbb{R}^{3} \xrightarrow{} \mathbf{c}.
\end{equation}

The design of our joint two-layer implicit model is inspired by the approach in~\cite{chen2022gdna}. However, as shown in Fig.~\ref{fig:implicit}, our model has two novel traits:
\emph{1) Instead of simply decomposing the $3$D clothed human into coarse and fine models, we apply a two-layer implicit model to represent the naked body and clothing shapes.
%
2) We additionally model texture to form a complete $3$D human model.}

\begin{figure}[t]
  \centering
  \includegraphics[width=0.88\linewidth]{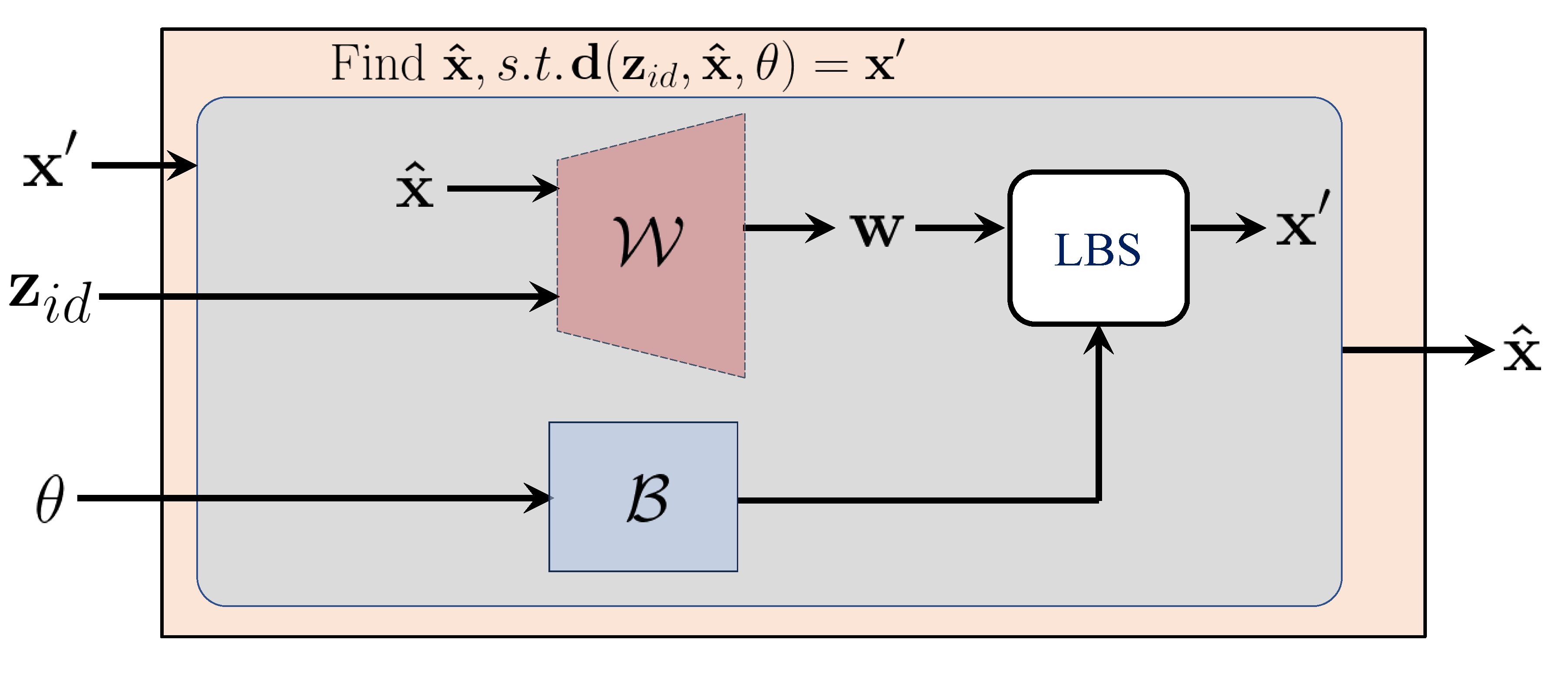}
   \caption{\small Neural blend skinning network. This module deforms the pose space to canonical space. Given a deformed point $\mathbf{x}'$, we compute its corresponding position $\mathbf{\hat{x}}$ in canonical space by iteratively finding the root of Eqn.~\ref{eqn:point_obj}.} 
   \label{fig:skinning}
\end{figure}

\subsection{Neural Linear Blend Skinning Network}

%
Our joint two-layer implicit model is built within the canonical space. However, the $3$D clothed human data, usually captured in various poses, introduces misalignments between this canonical space and the deformed counterpart.
%
%
To predict the occupancy values $\mathbf{o}_{1}$, $\mathbf{o}_{2}$ and the texture $\mathbf{c}$ for a given observed point $\mathbf{x}'$ within the deformed space, it is essential  to first determine its canonical correspondence point $\mathbf{\hat{x}}$. 
Once $\mathbf{x}$ is identified, we can then compute $\mathbf{o}_{1}$, $\mathbf{o}_{2}$ and $\mathbf{c}$ using Eqns.~\ref{eqn:naked_func},~\ref{eqn:clothed_func} and~\ref{eqn:texture_func}. 
The objective of this step is to find the canonical correspondence $\mathbf{\hat{x}}$ of any query point $\mathbf{x}'$.
To achieve this goal, similar to~\cite{chen2022gdna,chen2021snarf},  we learn 
a linear blend skinning (LBS)~\cite{loper2015smpl} using neural networks in an unsupervised manner.


\Paragraph{Regressing Blend Weight}
We follow~\cite{chen2021snarf,chen2022gdna} which define the skinning field in canonical space conditioned on our identity latent code $\mathbf{z}_{id}$:
\begin{align}
\label{eqn:weight}
\mathcal{W}:\mathbb{R}^{L_{id}}\times \mathbb{R}^3 &\xrightarrow{} \mathbb{R}^{K}  \nonumber\\
(\mathbf{z}_{id}, \mathbf{x}) &\xrightarrow{} \mathbf{w},
\end{align} 
where $\mathbf{w}$ is the point-wise blend weight of the canonical point $\mathbf{x}$. Then deformed point $\mathbf{x}'$ is determined by the following convex combination:
\begin{equation}
\label{eqn:trans}
\mathbf{x}' = \mathbf{d}(\mathbf{z}_{id}, \mathbf{x}, \mathbf{\theta}) = \sum_{k=1}^{K} \mathbf{w}_k\mathbf{B}_k\mathbf{x},
\end{equation} 
where $\mathbf{w}_{k}$ is $k$-element in the vector $\mathbf{w}$, while $\mathbf{B}_k$ denotes the $k$-element from the set of bone transformation matrices $\mathcal{B}=\{\mathbf{B}_{k}\in\mathbb{R}^{4\times 4}\}_{k=1}^{K}$.  
Both $\mathbf{x}$ and $\mathbf{x}'$ in Eqn.~\ref{eqn:trans} are represented in homogeneous coordinates. 

\Paragraph{Implicit Differentiable Skinning}
An overview is illustrated in Fig.~\ref{fig:skinning}. While the goal is to determine $\mathbf{x}'\xrightarrow{} \mathbf{\hat{x}}$, we only have direct access to the mapping defined by Eqn.~\ref{eqn:trans}, which is not invertible. 
%
%
Following~\cite{chen2021snarf,chen2022gdna}, the correspondence is calculated numerically by finding the root of the equation with Broyden's method~\cite{broyden1965class}:
\begin{equation}
\label{eqn:point_obj}
\mathbf{\hat{x}}= \{\mathbf{\hat{x}}| \mathbf{d}(\mathbf{z}_{id}, \mathbf{\hat{x}}, \mathbf{\theta}) - \mathbf{x}' = \mathbf{0}\}.
\end{equation}

\subsection{Implicit Rendering}\label{sec:render}
During rendering, when 2D pixels are unprojected  to $3$D points, they are intrinsically mapped in the deformed $3$D space.
Given a pixel $p$ of a masked input image, we construct a ray $\mathbf{x}'=\{\mathbf{c}_0+t\mathbf{v}|t\geqslant 0\}$, where $\mathbf{c}_0$ represents the camera's position and $\mathbf{v}$ indicates the viewing direction based on the camera projection parameters $\mathbf{\Omega}$. $t$ is the scalar distance along the ray.
We then map the ray points to the canonical space, following Eqn.~\ref{eqn:point_obj}.
The intersection point $\mathbf{\hat{x}}_p$ of the ray can be calculated by identifying the first change of clothed shape occupancy $o_2$.  
Finally, the rendered color of the pixel $p$ is calculated via Eqn.~\ref{eqn:texture_func}.

\subsection{Semi-supervised Model Learning}
While our model can perform self-supervised learning from real images without $3$D labels, we first pre-train our joint implicit $3$D model with $3$D data in order to mitigate the inherent ambiguity.

\subsubsection{Supervised Pre-training $3$D clothed Model}

\Paragraph{Training Data}
We combine CAPE~\cite{ma2020learning} ($3,000$ scans) and  THuman$2.0$~\cite{zheng2019deephuman} ($526$ scans) to train our joint two-layer implicit model. 
THuman$2.0$  consists of $526$ texture clothed $3$D scans of $105$ subjects. Following~\cite{chen2022gdna}, for each scan, we obtain its SMPL naked shape code. 
CAPE provides $148,584$ pairs of scans under clothing and SMPL naked body with rich pose variations of $15$ subjects. 
We randomly sample $3,000$ scans for training. 
Formally, each training sample can be represented as SMPL pose $\mathbf{\theta}$, identity label $l_{3D}$, $n$ spatial points $\mathbf{x}'_i$, and their SDFs $o^i_{1}, o^i_{2}$, and color $\mathbf{c}_i$: $\{ \theta, l_{3D}, \{\mathbf{x}'_i, o^i_{1}, o^i_{2}, \mathbf{c}_i\}_{i=1}^{n}\}$.
With the autodecoding technique~\cite{park2019deepsdf}, we  assign trainable identity shape code, clothing shape code, and texture code to each training sample.

\Paragraph{Loss Function} 
We define the loss below for each sample:
\vspace{-2mm}
\begin{gather}
\argmin_{\mathcal{F}, \mathcal{C}, \mathcal{T}, \mathbf{z}_{id}, \mathbf{z}_{cloth}, \mathbf{z}_{tex}} \mathcal{L}_{id} + \mathcal{L}_{cloth} + \mathcal{L}_{tex} +\mathcal{L}^{3D}_{cla}(\mathbf{z}_{id}, l_{3D}) \nonumber  \\
\mathcal{L}_{id}={\textstyle\sum}_{i=0}^{n}BCE(\mathcal{F}(\mathbf{z}_{id}, \mathbf{\hat{x}}_i), o^{i}_{1}) \\
\mathcal{L}_{cloth}={\textstyle\sum}_{i=0}^{n}BCE(\mathcal{C}(\mathbf{z}_{cloth},\mathbf{f}_i,\mathbf{\hat{x}}_i), o^{i}_{2}) \\
\mathcal{L}_{tex}={\textstyle\sum}_{i=0}^{n}||\mathcal{T}(\mathbf{z}_{tex},\mathbf{z}_{cloth},\mathbf{\hat{x}}_i)-\mathbf{c}_{i}||_2,  
\end{gather} 
where $\mathcal{L}^{3D}_{cla}(\mathbf{z}_{id}, l_{3D})$ is the cross-entropy classification loss.
We additionally add auxiliary loss $\mathcal{L}_{W}$ to train network $\mathcal{W}$:
\vspace{-2mm}
\begin{gather}
\label{eqn:auxiliary_loss}
{\textstyle\argmin}_{\mathcal{W}}  \mathcal{L}_{W}  \\
\mathcal{L}_{W}={\textstyle\sum}_{k=0}^{K}||\mathcal{W}(\mathbf{z}_{id},\mathbf{J}_k)-\mathbf{w}_{\mathbf{J}_k}||_2,  
\end{gather} 
where 
$\mathbf{w}_{\mathbf{J}_k}$ is the pre-computed ground truth skinning weights of SMPL joints location $\mathbf{J}_k$. 

\subsubsection{Self-supervised Joint Modeling and Fitting}
Given a set of in-the-wild $2$D images with body masks and identity labels $\{\mathbf{I}_{i},\mathbf{M}_i, l_{i}\}_{i=1}^{T}$, the self-supervised identity disentanglement loss is:
\vspace{-1mm}
\begin{equation}
\label{eqn:self_sup_loss}
\argmin_{\mathcal{E}, \mathcal{T}} {\textstyle\sum}_{i=1}^{T}\mathcal{L}_{sil} + \mathcal{L}_{rgb} + \mathcal{L}_{cla},
\end{equation} 
where $\mathcal{L}_{rgb} $ is the photometric loss, $\mathcal{L}_{sil} $ is silhouette loss and $\mathcal{L}_{cla}$ is the classification loss. Specifically,
we denote $\mathbf{I}_{p}$ and $\mathbf{M}_{p}$ as the RGB and silhouette values of pixel $p\in P$.
Here, $P$ denotes the entire set of pixels in the input image $\mathbf{I}$. A subset of $P$, represented as $P^{in}$, corresponds to the pixels where an intersection between the rays and the body in the image has been detected. 
%
The photometric loss is defined as
\vspace{-1mm}
\begin{equation}
\label{eqn:rgb_loss}
\mathcal{L}_{rgb} = \frac{1}{|P|}\sum_{p\in P^{in}}|\mathbf{I}_p-\mathcal{T}(\mathcal{E}_{tex}(\mathbf{I}), \mathcal{E}_{cloth}(\mathbf{I}), \mathbf{\hat{x}}_p)|,
\end{equation} 
where the encoder $\mathcal{E}$ estimates $\mathbf{z}_{id}=\mathcal{E}_{id}(\mathbf{I})$, $\mathbf{z}_{cloth}=\mathcal{E}_{cloth}(\mathbf{I})$ and $\mathbf{z}_{tex}=\mathcal{E}_{tex}(\mathbf{I})$ from image $\mathbf{I}$. $\mathbf{\hat{x}}$ is the intersection point (see Sec.~\ref{sec:render}).  

We further define the silhouette loss as
\begin{equation}
\label{eqn:sil_loss}
\mathcal{L}_{sil} = \frac{1}{|P|}\sum_{p\in P^{out}}CE(\mathbf{M}_p, \mathbf{\hat{M}}_p),
\end{equation} 
where $\mathbf{\hat{M}}$ is the masked rendering, $P^{out}=P - P^{in}$ represents the indices in the mini-batch for which there is no ray-geometry intersection or $\mathbf{M}_p=0$, and $CE(\cdot,\cdot)$ denotes the cross-entropy loss.
We impose triplet loss and cross-entropy loss on the identity shape code $\mathbf{z}_{id}=\mathcal{E}_{id}(\mathbf{I})$ as our classification loss $ \mathcal{L}_{cla}(\mathbf{z}_{id}, l)$.

\begin{figure}[t]
\begin{center}
\resizebox{1\linewidth}{!}{
\begin{tabular}{c c}
\red Challenging set & \blue Normal set \\
\red{\includegraphics[scale=0.37]{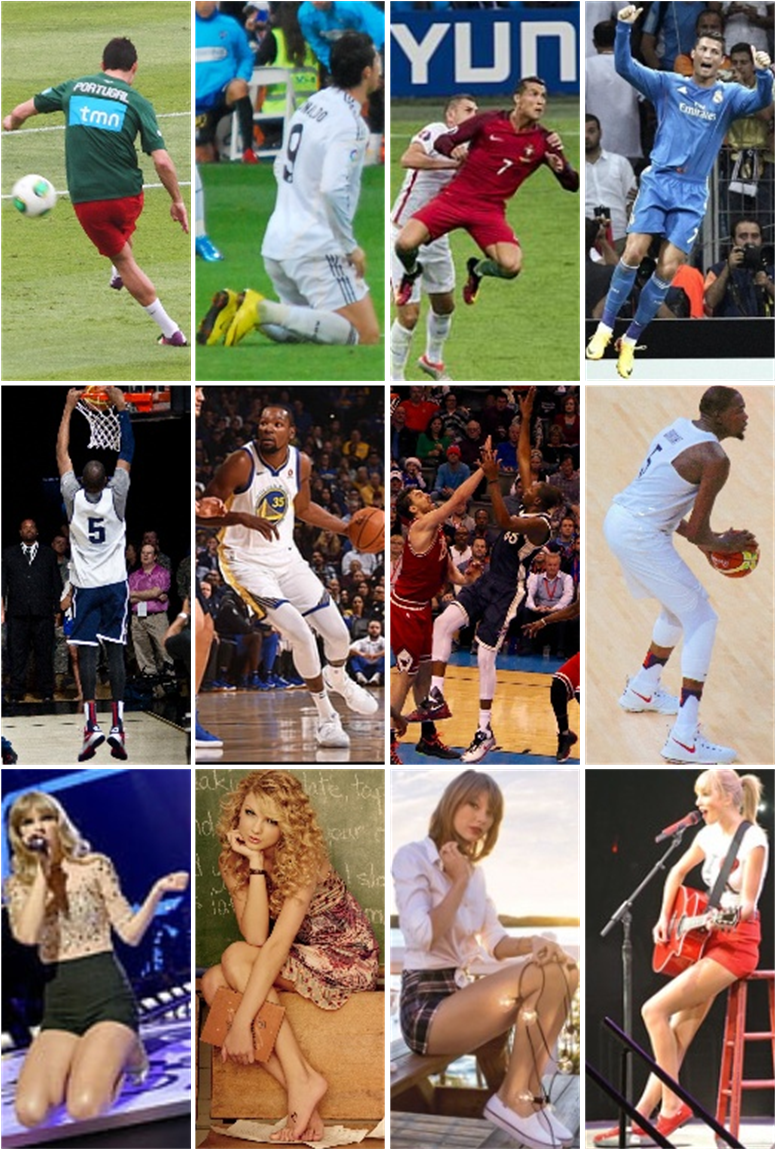}} & \hspace{-4mm} \blue{\includegraphics[scale=0.37]{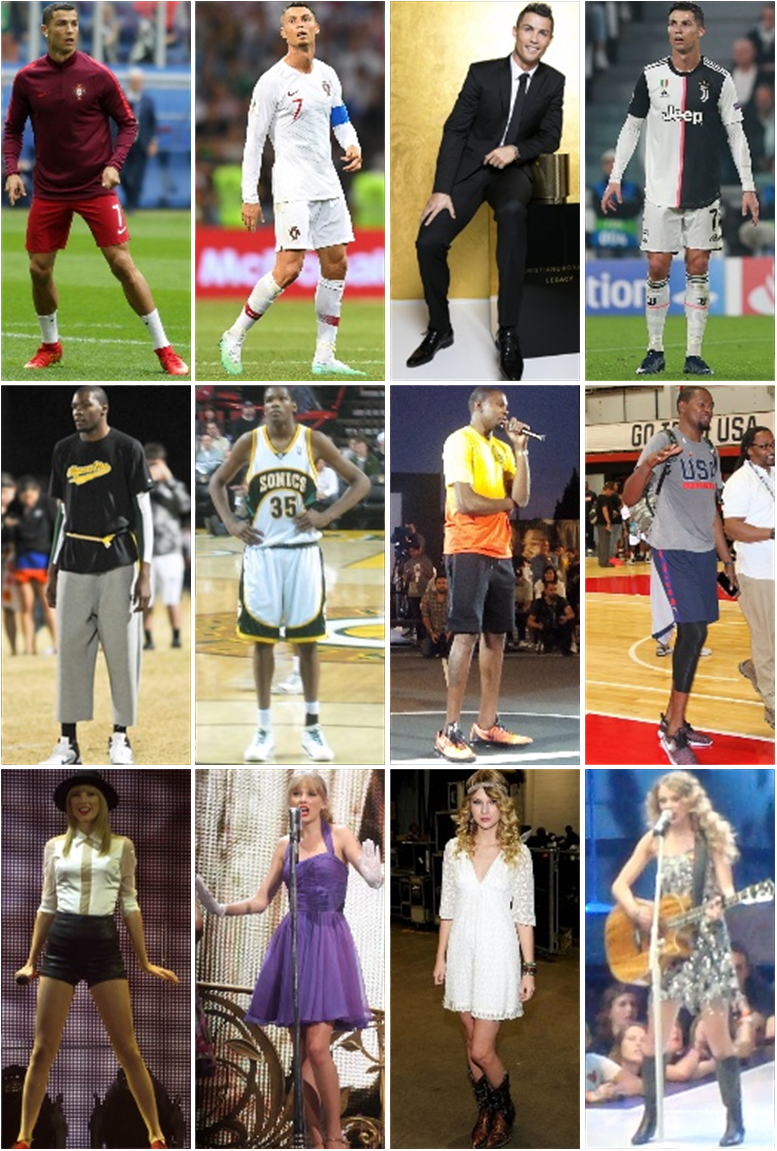}} \\
\end{tabular}
}
\vspace{0mm}
\caption{\small Example images from CCDA, with one subject per row, showcasing the diversity of body poses, clothing styles and colors.
} 
\label{fig:dataset_exam}
\end{center}
\vspace{-4mm}
\end{figure}

\subsection{Person ReID Inference}
For person ReID inference, the encoder $\mathcal{E}$ processes body images and extracts the identity shape features $\mathbf{z}_{id}$. The Cosine similarity of $\mathbf{z}_{id}$ is then used to determine if the two images belong to the same person. It is worth noting that, the inference of our ReID does not incorporate the $3$D reconstruction module, making it highly efficient.

\section{CCDA Person ReID Dataset}\label{sec:dataset}
We construct a new dataset with diverse human  activities and clothing changes for evaluating LT-ReID. 
%
Specifically, we collect data for popular athletes in soccer, tennis, and basketball, and popular artists, such as fashion models and singers. We crawl whole body images of each subject on Google Image\footnote{ All collected images are under \texttt{Creative Commons licenses}.} with athlete/artist names. 
We collect two sets of images per subject: `challenging' and `normal' body poses. 
As an example, for a basketball player, the `challenging' set includes images of players' actions on the court, while the `normal' set contains standing or walking poses. 
%
We then crop the body region from the original image via the detected bounding box and resize it to $256\times128$. 
Finally, the annotator verifies the identity of each image. 
In total, $1,555$ images of $100$ subjects are retained. For each subject, we randomly select one image from `normal' images for the gallery set, while the remaining $1,455$ images comprise the query set.
%
%
Fig.~\ref{fig:dataset_exam} shows examples of images in the CCDA dataset. 

%% file: sec_4_exp.tex
\Section{Experimental Results}\label{sec:exp}

\Paragraph{Implementation Details}
Our training process includes two stages: 1) Networks $\mathcal{F}$, $\mathcal{C}$, $\mathcal{T}$, $\mathcal{W}$ are pre-trained on $3$D data. 2) $\mathcal{E}$ and $\mathcal{T}$ are trained or fine-tuned with real images. The encoder $\mathcal{E}$ is implemented as a ResNet-$50$. Networks $\mathcal{F}$, $\mathcal{C}$, $\mathcal{T}$ and $\mathcal{W}$ are MLPs. In experiments, we set $L_{cloth}=L_{tex}=512$, $L_{id}=4,096$, $L_{f}=L_{h}=256$, $K=24$, $n=200,000$. We implement in Pytorch and use Adam optimizer in both stages. 

\subsection{Person ReID}

\Paragraph{Metric} For person ReID, we follow the standard retrieval accuracy metrics, namely the Cumulative Matching Characteristics (CMC) and mean average precision (mAP). 

\Paragraph{Baseline} We compare our method with eight SoTA person ReID methods: Two-Stream~\cite{zheng2017discriminatively}, MLFN~\cite{chang2018multi}, HACNN~\cite{li2018harmonious}, Part-Aligned~\cite{suh2018part}, PCB~\cite{sun2018beyond}, TriNet~\cite{hermans2017defense}, MGN~\cite{wang2018learning}, DG-Net~\cite{zheng2019joint}, and five SoTA cloth-changing re-ID methods: ReIDCaps~\cite{huang2019beyond}, 
$3$DSL~\cite{chen2021learning}, RCSAnet~\cite{huang2021clothing}, FSAM~\cite{hong2021fine} and CAL~\cite{gu2022clothes}.

\subsubsection{Results on Cloth-changing Person ReID datasets}

\Paragraph{Datasets} We test on four popular cloth-changing  ReID datasets: Celeb-reID/Celeb-reID-light~\cite{huang2019beyond,huang2019celebrities}, PRCC~\cite{yang2019person}, LTCC~\cite{shu2021large} and the recent CCVID dataset~\cite{gu2022clothes,zhang2019gait}.



\begin{table}[t!]
\newcommand{\tabincell}[2]{\begin{tabular}{@{}#1@{}}#2\end{tabular}}
\newcommand{\greencheck}{\textcolor{green}{\ding{51}}} 
\newcommand{\redX}{\textcolor{red}{\ding{55}}}
\newcommand{\halfcheck}{\textcolor{green}{\ding{51}}{\small\textcolor{black}{\msquare }}}
\begin{center}
\resizebox{1\linewidth}{!}{
\begin{tabular}{r |c| c | c | c | c  }
\hline
\multirow{2}{*}{Method} &  \multirow{2}{*}{ Backbone}   & \multicolumn{2}{c}{Celeb-reID }  & \multicolumn{2}{|c}{Celeb-reID-light }   \\
\cline{3-6}
&   & mAP & Rank$1$  & mAP & Rank$1$  \\
\hline\hline
Two-Stream~\cite{zheng2017discriminatively}    &  ResNet-$50$ & $7.8$  & $36.3$ & - & - \\
MLFN~\cite{chang2018multi}          & * & $6.0$  & $41.4$ & $6.3$  & $10.6$ \\
HACNN~\cite{li2018harmonious}         & * & $9.5$  & $47.6$ & $11.5$ & $16.2$ \\
Part-Aligned~\cite{suh2018part}  & GoogLeNet\textcolor{red}{\textbf{$\times2$}} & $6.4$  & $19.4$ & - & - \\
PCB~\cite{sun2018beyond}           & ResNet-$50$ & $8.2$  & $37.1$ & - & - \\
MGN~\cite{wang2018learning}           &  ResNet-$50$  & $10.8$ & $49.0$ & $13.9$ & $21.5$ \\
DG-Net~\cite{zheng2019joint}        &  ResNet-$50$ & $10.6$ & $50.1$ & $12.6$ & $23.5$ \\
\hline
\multicolumn{6}{c}{\gray  \emph{cloth-changing person ReID methods}} \\
\hdashline
ReIDCaps-~\cite{huang2019beyond}    & DenseNet-$121$ & $9.8$  & $51.2$ & $11.2$ & $20.3$ \\
ReIDCaps~\cite{huang2019beyond}   & DenseNet-$121$\textcolor{red}{\textbf{$\times6$}} & $15.8$ & $63.0$ & $19.0$ & $33.5$ \\
RCSAnet~\cite{huang2021clothing}     & DenseNet-$121$\textcolor{red}{\textbf{$\times2$}} & $11.9$ & $55.6$ & $16.7$ & $29.5$ \\
CAL~\cite{gu2022clothes}& ResNet-$50$ & $1
3.7$ & $59.2$ & $18.5$ & $33.6$ \\ 
\hline
\rowcolor{LightCyan}
\textbf{3DInvarReID}   & ResNet-$50$ & $11.8$ & $55.2$ & $15.0$ & $30.1$ \\  
\rowcolor{LightCyan}
\textbf{3DInvarReID$^\#$}   & ResNet-$50$ & $15.2$ & $61.2$ & $21.8$ & $37.0$ \\ [-0.3ex] 
\rowcolor{rowred}
\multirow{2}{*}{ReIDCaps+} & \multirow{2}{*}{-} & \multirow{2}{*}{$\textbf{18.4}$} & \multirow{2}{*}{$\textbf{65.5}$} & \multirow{2}{*}{$\textbf{25.7}$} & \multirow{2}{*}{$\textbf{42.2}$} \\[1.0ex] 
\rowcolor{rowred}
\textbf{3DInvarReID$^\#$} & & & & & \\ [-0.3ex] 


\hline
\end{tabular}
}
\end{center}
\vspace{0mm}
\caption{\small Comparison with SoTA on Celeb-reID and Celeb-reID-light datasets (\%). `*' indicates that the backbone is designed by the authors. The red number means the total number of models. \textbf{3DInvarReID$^\#$}'s weights are initialized using the CAL model.}
\label{tab:celeb}
\end{table}

\Paragraph{Results on Celeb-reID and Celeb-reID-light}
As reported in Tab.~\ref{tab:celeb}, our \textbf{3DInvarReID} (\textbf{Ours}) outperforms all the general person re-ID baselines on both datasets. 
Considering methods using a {\it single} network, we also outperform the cloth-changing baseline, ReIDCaps-~\cite{huang2019beyond}.  
More importantly, to investigate the complementarity between our learned $3$D shape features and existing $2$D features, we fuse our method with cloth-changing baselines by simple summation at the score level. 
By fusing with ReIDCaps~\cite{huang2019beyond}, our method improves the Rank$1$ accuracy on Celeb-reID from $63.0\%$ to $65.5\%$ and from $33.5\%$ to $42.2\%$ on Celeb-reID-light. 
These results clearly demonstrate that the $3$D shape features learned from our method are both discriminative and complementary to the $2$D features, indicating the effectiveness of our proposed approach for person ReID, particularly under cloth-changing scenarios.

%

\begin{table}[t!]
\newcommand{\tabincell}[2]{\begin{tabular}{@{}#1@{}}#2\end{tabular}}
\newcommand{\greencheck}{\textcolor{green}{\ding{51}}} 
\newcommand{\redX}{\textcolor{red}{\ding{55}}}
\newcommand{\halfcheck}{\textcolor{green}{\ding{51}}{\small\textcolor{black}{\msquare }}}
\begin{center}
\resizebox{0.95\linewidth}{!}{
\begin{tabular}{r |c| c | c | c | c  }
\hline
\multirow{2}{*}{Method} &  \multirow{2}{*}{ Backbone}   & \multicolumn{2}{c}{LTCC }  & \multicolumn{2}{|c}{PRCC }   \\
\cline{3-6}
&   & mAP & Rank$1$  & mAP & Rank$1$  \\
\hline\hline
HACNN~\cite{li2018harmonious}     & * & $9.3$  & $21.6$ & $-$ & $21.8$ \\
PCB~\cite{sun2018beyond}     & ResNet-$50$ & $10.0$  & $23.5$ & $38.7$ & $41.8$ \\
$3$DSL~\cite{chen2021learning}   & ResNet-$50$\textcolor{red}{\textbf{$\times2$}} & $14.8$  & $31.2$ & $-$ & $51.3$ \\
FSAM~\cite{hong2021fine}    & ResNet-$50$ & $16.2$  & $38.5$ & $-$ & $54.5$ \\
CAL~\cite{gu2022clothes}   & ResNet-$50$ & $18.0$ & $40.1$ & $55.8$ & $55.2$ \\
\hline
\rowcolor{LightCyan}
\textbf{3DInvarReID}   & ResNet-$50$ & $16.7$ & $37.8$ & $52.5$ & $51.6$ \\ [-0.3ex]
\rowcolor{rowred}
\multirow{2}{*}{CAL+} & \multirow{2}{*}{-} & \multirow{2}{*}{$\textbf{18.9}$} & \multirow{2}{*}{$\textbf{40.9}$} & \multirow{2}{*}{$\textbf{57.2}$} & \multirow{2}{*}{$\textbf{56.5}$} \\[1.0ex]
\rowcolor{rowred}
\textbf{3DInvarReID} & & & & & \\ \hline

\rowcolor{rowlightblue} \emph{CAL}   & ResNet-$50$ & $2.8$ & $3.8$ & $20.3$ & $31.6$ \\  
\rowcolor{rowblue}  \emph{\textbf{3DInvarReID}}    & ResNet-$50$  & $2.8$ & $5.1$ & $20.1$ & $34.6$ \\
\rowcolor{rowblue}  \emph{\textbf{3DInvarReID}$^\#$}    & ResNet-$50$  & $\textbf{2.8}$ & $\textbf{5.6}$ & $\textbf{21.4}$ & $\textbf{40.7}$ \\
\hline

\end{tabular}
}
\end{center}
\vspace{0mm}
\caption{\small Comparison with SoTA cloth-changing person ReID methods on the LTCC and PRCC datasets (\%). Models highlighted in pink are trained on the Celeb-reID dataset. \textbf{3DInvarReID$^\#$}'s weights are initialized using the CAL model.}
\label{tab:ltcc_prcc}
\vspace{0mm}
\end{table}

\begin{table}[t!]
\newcommand{\tabincell}[2]{\begin{tabular}{@{}#1@{}}#2\end{tabular}}
\newcommand{\greencheck}{\textcolor{green}{\ding{51}}} 
\newcommand{\redX}{\textcolor{red}{\ding{55}}}
\newcommand{\halfcheck}{\textcolor{green}{\ding{51}}{\small\textcolor{black}{\msquare }}}
\begin{center}
\resizebox{0.95\linewidth}{!}{
\begin{tabular}{r |c| c | c | c | c  }
\hline
\multirow{2}{*}{Method} &  \multirow{2}{*}{ Backbone}   & \multicolumn{2}{c}{General }  & \multicolumn{2}{|c}{Cloth-changing }   \\
\cline{3-6}
&   & mAP & Rank$1$  & mAP & Rank$1$  \\
\hline\hline

TriNet~\cite{hermans2017defense}    & ResNet-$50$ & $78.1$  & $81.5$ & $77.0$ & $81.1$ \\
CAL~\cite{gu2022clothes}   & ResNet-$50$ & $81.3$ & $82.6$ & $79.6$ & $81.7$ \\
\hline
\rowcolor{LightCyan}
\textbf{3DInvarReID}   & ResNet-$50$ & $66.1$ & $70.8$ & $65.4$ & $70.2$ \\ [-0.3ex]
\rowcolor{rowred}
\multirow{2}{*}{CAL+} & \multirow{2}{*}{-} & \multirow{2}{*}{$\textbf{82.6}$} & \multirow{2}{*}{$\textbf{83.9}$} & \multirow{2}{*}{$\textbf{81.3}$} & \multirow{2}{*}{$\textbf{84.3}$} \\[1.0ex]
\rowcolor{rowred}
\textbf{3DInvarReID} & & & & & \\
\hline
\end{tabular}
}
\vspace{0mm}
\end{center}
\caption{\small Comparison with SoTA methods on CCVID (\%).}
\label{tab:ccvid}
\vspace{0mm}
\end{table}
\begin{table}[t!]
\newcommand{\tabincell}[2]{\begin{tabular}{@{}#1@{}}#2\end{tabular}}
\newcommand{\greencheck}{\textcolor{green}{\ding{51}}} 
\newcommand{\redX}{\textcolor{red}{\ding{55}}}
\newcommand{\halfcheck}{\textcolor{green}{\ding{51}}{\small\textcolor{black}{\msquare }}}
\begin{center}
\resizebox{0.9\linewidth}{!}{
\begin{tabular}{r |c| c | c | c  }
\hline
Method &  Backbone  & mAP & Rank$1$  & Rank$5$ \\
\hline\hline

ReIDCaps~\cite{huang2019beyond}   & DenseNet-$121$\textcolor{red}{\textbf{$\times6$}} & $10.9$ & $6.5$ & $20.2$  \\
CAL~\cite{gu2022clothes}     & ResNet-$50$ & $19.3$ & $10.0$ & $26.7$ \\ \hline  
\rowcolor{rowred}  \textbf{3DInvarReID}  & ResNet-$50$ & $\textbf{21.7}$ & $\textbf{11.1}$  & $\textbf{30.5}$\\
\hline
\end{tabular}
}
\end{center}
\vspace{0mm}
\caption{\small Comparison with SoTA methods on CCDA dataset (\%).}
\label{tab:ccda}
\vspace{0mm}
\end{table}

\Paragraph{Results on LTCC, PRCC and CCVID}
The comparison of the LTCC, PRCC and CCVID datasets is shown in Tabs.~\ref{tab:ltcc_prcc} and \ref{tab:ccvid}. 
Similarly, by fusing the best baseline CAL~\cite{gu2022clothes}, our method achieves additional improvements.
For instance, when evaluated in the cloth-changing setting, our method achieves a significant improvement of $2.6\%$ in Rank$1$ accuracy by fusing with CAL on the CCVID dataset. 
These findings highlight the effectiveness of \textbf{3DInvarReID}, with the $3$D shape features being shown to be both discriminative and complementary to $2$D features. 
Our approach stands out for its superior performance compared to other $3$D feature extraction methods for person ReID. This is particularly noteworthy in comparison to the 3DSL~\cite{chen2021learning} (Tab.~\ref{tab:ltcc_prcc}).
%
Additionally, we assess our model under a \emph{\textbf{cross-domain setting}}—that is, the model is trained with the Celeb-reID dataset and tested using the LTCC and PRCC datasets.
Table~\ref{tab:ltcc_prcc} shows that our models outperform the baseline CAL, indicating a superior discriminative feature representation.


\subsubsection{Results on LT-ReID dataset (CCDA)}
Given that both our CCDA and Celeb-reID datasets are obtained from the Internet and share a similar image style, we choose trained models on Celeb-reID and evaluate them on CCDA. We choose the SoTA cloth-changing methods, ReIDCaps-~\cite{huang2019beyond} and CAL~\cite{gu2022clothes} as baselines. 
The results in Tab.~\ref{tab:ccda} demonstrate that our \textbf{3DInvarReID} outperforms the baselines, providing strong evidence of its effectiveness in handling person ReID with challenging body poses and cloth-changing variations.


\subsubsection{Results on Short-term Person ReID datasets}

Despite our method being designed for LT-ReID, we additionally compare with SoTA methods on two conventional short-term ReID datasets: Market-1501~\cite{zheng2015scalable} and MSMT17~\cite{wei2018person}, in Tab.~\ref{tab:market_msmt}. By fusing with CAL~\cite{gu2022clothes}, we observe an average improvement of $2.7\%$ in Rank$1$ accuracy on both datasets, demonstrating the complementary nature of our 3D shape feature, even on short-term datasets.

\begin{table}[t!]
\newcommand{\tabincell}[2]{\begin{tabular}{@{}#1@{}}#2\end{tabular}}
\newcommand{\greencheck}{\textcolor{green}{\ding{51}}} 
\newcommand{\redX}{\textcolor{red}{\ding{55}}}
\newcommand{\halfcheck}{\textcolor{green}{\ding{51}}{\small\textcolor{black}{\msquare }}}
\begin{center}
\resizebox{0.95\linewidth}{!}{
\begin{tabular}{r |c| c | c | c | c  }
\hline
\multirow{2}{*}{Method} &  \multirow{2}{*}{ Backbone}   & \multicolumn{2}{c}{Market-1501}  & \multicolumn{2}{|c}{MSMT17}   \\
\cline{3-6}
&   & mAP & Rank$1$  & mAP & Rank$1$  \\
\hline\hline
PCB~\cite{sun2018beyond}     & ResNet-$50$ & $81.6$  & $93.8$ & $40.4$ & $68.2$ \\
$3$DSL~\cite{chen2021learning}   & ResNet-$50$\textcolor{red}{\textbf{$\times2$}} & $87.3$  & $95.0$ & $-$ & $-$ \\
FSAM~\cite{hong2021fine}    & ResNet-$50$ & $85.6$  & $94.6$ & $-$ & $-$ \\
CAL~\cite{gu2022clothes}   & ResNet-$50$ & $87.5$ & $94.7$ & $57.3$ & $79.7$ \\
\hline
\textbf{3DInvarReID}   & ResNet-$50$ & $85.5$ & $94.2$ & $55.1$ & $76.3$ \\[-0.3ex]
\rowcolor{rowred}
\multirow{2}{*}{CAL+} & \multirow{2}{*}{-} & \multirow{2}{*}{$\textbf{87.9}$} & \multirow{2}{*}{$\textbf{95.1}$} & \multirow{2}{*}{$\textbf{59.1}$} & \multirow{2}{*}{$\textbf{80.8}$} \\[1.0ex]
\rowcolor{rowred}
\textbf{3DInvarReID} & & & & & \\

\hline

\end{tabular}
}
\vspace{0mm}
\end{center}
\caption{\small Comparison on short-term ReID  datasets (\%).}
\label{tab:market_msmt}
\vspace{0mm}
\end{table}

\subsection{3D Reconstruction}
\begin{figure}[t]
\begin{center}
\begin{tabular}{c c c c c}
\vspace{-2mm}
\small Input & \hspace{-12mm} \small  LVD~\cite{corona2022learned} & \hspace{-12mm} \small  SHAPY~\cite{choutas2022accurate} & \hspace{-12mm} \small  Ours &  \hspace{-12mm} \small  GT \\ 
\hspace{-1mm} {\includegraphics[height=10cm]{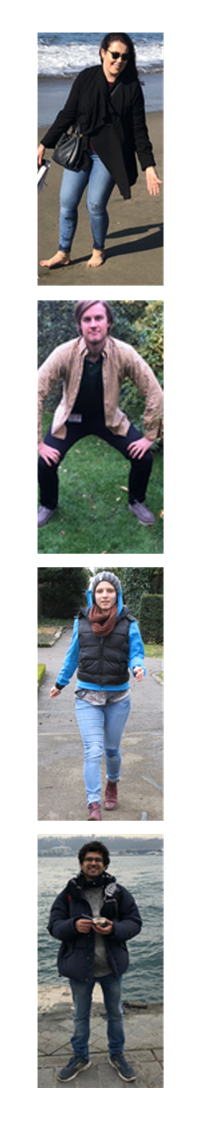}} & \hspace{-12mm}
{\includegraphics[height=10cm]{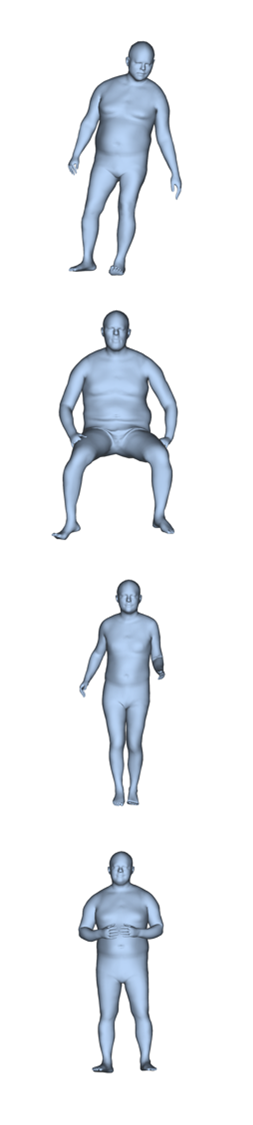}} & \hspace{-12mm}
{\includegraphics[height=10cm]{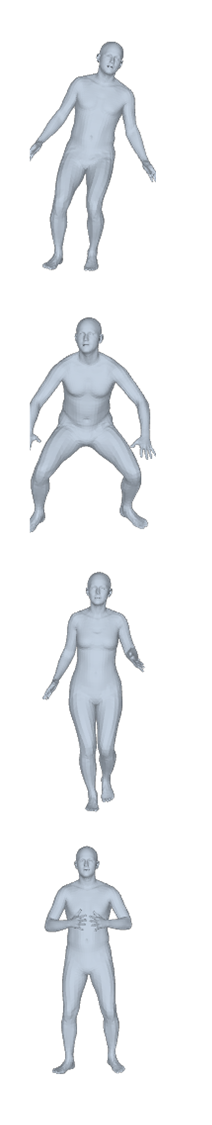}} & \hspace{-12mm}
{\includegraphics[height=10cm]{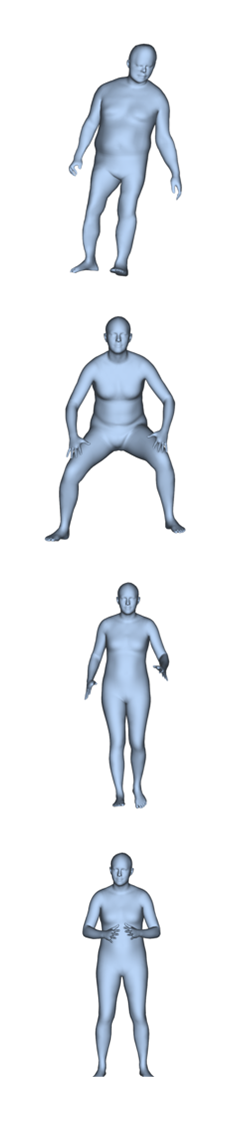}} & \hspace{-12mm}
{\includegraphics[height=10cm]{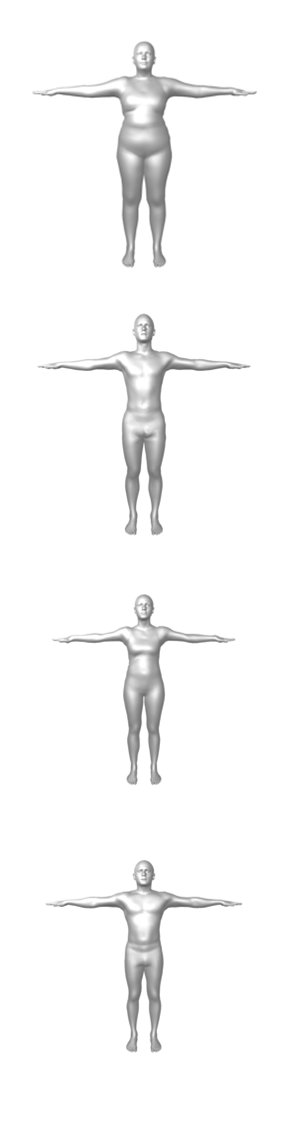}} \\
\end{tabular}
\vspace{-3mm}
   \caption{\small  Qualitative comparisons with LVD~\cite{corona2022learned} and SHAPY~\cite{choutas2022accurate} on $3$D naked body reconstruction. Our approach recovers more accurate $3$D body shapes from the images. \vspace{-5mm}} 
\label{fig:3d_performance1}
\end{center}
\vspace{0mm}
\end{figure}

\begin{figure}[t]
\begin{center}
\resizebox{1\linewidth}{!}{
\begin{tabular}{c c c c}
\vspace{-2mm}
\small Input & \hspace{-6mm} \small  ICON~\cite{xiu2022icon} & \hspace{-6mm} \small  ClothWild~\cite{moon20223d} &  \hspace{-6mm} \small  Ours \\ 
{\includegraphics[height=8cm]{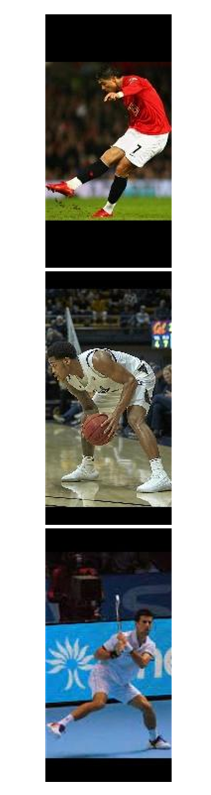}} & \hspace{-6mm}
{\includegraphics[height=8cm]{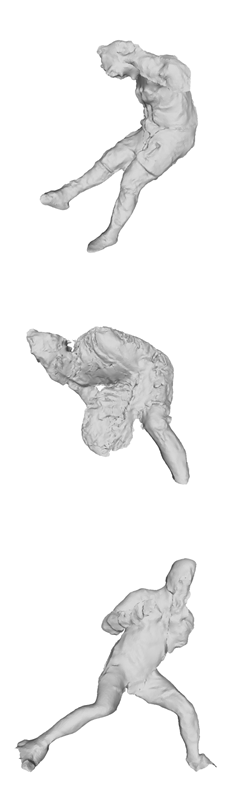}} & \hspace{-6mm}
{\includegraphics[height=8cm]{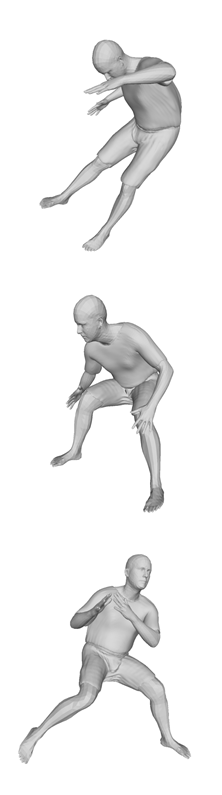}} & \hspace{-6mm}
{\includegraphics[height=8cm]{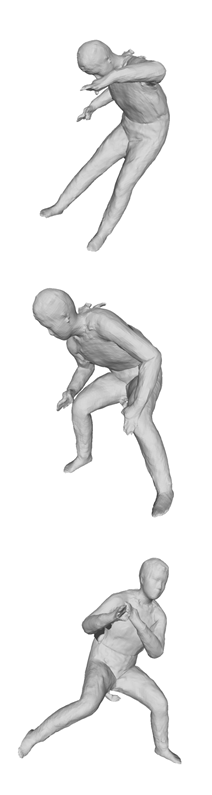}} \\
\end{tabular}
}
\vspace{0mm}
   \caption{\small Qualitative  comparisons with ICON~\cite{xiu2022icon} and ClothWild~\cite{moon20223d} on $3$D clothed body reconstruction.} 
\label{fig:3d_performance2}
\end{center}
\vspace{-3mm}
\end{figure}

Most $3$D body reconstruction methods focus more on pose estimation than shape estimation. 
Recently, SHAPY~\cite{choutas2022accurate} releases a dataset (HBW) that contains ground-truth 3D body scans and the corresponding in-the-wild images, which enables us to test the accuracy of our reconstructed 3D naked body shapes.
We thus evaluate our methods on the validation set of HBW, which contains $237$ in-the-wild images of $10$ subjects. 
Our baseline includes LVD~\cite{corona2022learned} and SHAPY~\cite{choutas2022accurate}, which are recent pipelines for discriminative identity shape fitting.
Following~\cite{corona2022learned}, we evaluate the reconstruction accuracy with Chamfer distance (CD-$L_2$), by uniformly sampling $10,000$ points on both ground-truth and predicted meshes in the canonical space.
As shown in Tab.~\ref{tab:recon}, our method outperforms the two baselines. We also visualize $3$D reconstructions in Fig.~\ref{fig:3d_performance1}. Our reconstructions resemble the ground truth better than the baselines. These results demonstrate the superiority of the proposed method in reconstructing naked $3$D body shapes.
Fig.~\ref{fig:3d_performance2} shows qualitative comparisons with  ICON~\cite{xiu2022icon} and ClothWild~\cite{moon20223d}. 
Our approach achieves comparable clothed $3$D body reconstructions. 

\begin{table}[t!]
\vspace{0mm}
\begin{center}
\resizebox{0.6\linewidth}{!}{
\begin{tabular}{l |c| c | c   }
\hline
  & LVD~\cite{corona2022learned} &SHAPY~\cite{choutas2022accurate}  & Ours \\
\hline\hline
CD-$L_2$ & $0.654$ & $0.632$ & $\textbf{0.610}$ \\

\hline
\end{tabular}
}
\end{center}
\vspace{0mm}
\caption{\small Comparison of $3$D body reconstruction on HBW.}
\label{tab:recon}
\vspace{0mm}
\end{table}

\begin{table}[t!]
\vspace{0mm}
\newcommand{\tabincell}[2]{\begin{tabular}{@{}#1@{}}#2\end{tabular}}
\newcommand{\greencheck}{\textcolor{green}{\ding{51}}} 
\newcommand{\redX}{\textcolor{red}{\ding{55}}}
\newcommand{\halfcheck}{\textcolor{green}{\ding{51}}{\small\textcolor{black}{\msquare }}}
\begin{center}
\resizebox{0.9\linewidth}{!}{
\begin{tabular}{l |c| c | c   }
\hline
 Model Type  & mAP & Rank$1$  & Rank$5$ \\
\hline\hline

\rowcolor{LightCyan}  \textbf{3DInvarReID} &
$\textbf{17.5}$ & $\textbf{36.3}$  & $\textbf{56.4}$\\
\textbf{3DInvarReID}-w/o $3$D  & $14.6$ & $28.4$  & $ 48.3$\\
\textbf{3DInvarReID}-w/o Pre-training  & $15.1$ & $29.3$  & $ 50.7$\\
\textbf{3DInvarReID}-w/o $3$D clothing   & $16.0$ & $32.1$  & $ 52.8$\\

\hline
\end{tabular}
}
\end{center}
\vspace{0mm}
\caption{\small Ablation studies on CCDA dataset (\%).}
\label{tab:ablation}
\vspace{0mm}
\end{table}

\subsection{Ablation Study}
In this section, all models are trained on Celeb-reID and tested on the CCDA dataset.

\Paragraph{Effect of the $3$D module} 
We compare our full model with an ablated version that only incorporates the $\mathcal{L}_{cla}$ loss in its training,  disregarding $3$D modules.
The results in Tab.~\ref{tab:ablation} show that our \textbf{3DInvarReID} significantly improves the recognition accuracy, leading to a Rank$1$ accuracy increase from $28.4$ to $36.3$.


\Paragraph{Effect of Our Two-layer Implicit Model} 
Our primary goal is to disentangle the identity feature from non-identity features in $3$D shape space. To evaluate the effectiveness of our $3$D disentanglement module, we train a model (\textbf{Ours}-w/o $3$D clothing) by replacing our $3$D body model with SMPL shape bases and omitting the modeling of the $3$D clothing component and rendering layer. The results in Tab.~\ref{tab:ablation} demonstrate the advantages of modeling clothing shape and texture for person ReID (Rank$1$: $32.1$$\xrightarrow{}$$36.3$).

\Paragraph{Effect of Pre-training} 
Tab.~\ref{tab:ablation} shows that removing the pre-training stage results in a lower Rank$1$ accuracy, with a score of $29.3$ compared to our $36.3$. This finding highlights the effectiveness of pre-training in addressing the inherent ambiguity of disentanglement.



%% file: sec_5_conclusion.tex
\section{Conclusions}\label{sec:con}

This paper tackles the challenging setting of long-term person ReID, which allows a wider range of real-world human activities and accounts for cloth-changing scenarios.
To address this problem, we present a joint two-layer implicit representation to model textured $3$D clothed humans together with a discriminative fitting module, enabling us to disentangle identity and non-identity features for real-world images. 
We collect a new LT-ReID dataset, CCDA, with diverse human activities and clothing changes, facilitating future research on real-world scenarios.
Experimental results demonstrate the effectiveness of our method in disentangling identity and non-identity features in $3$D clothed body shapes, thereby contributing to LT-ReID.

\Paragraph{Limitations \& Potential Negative Impacts Impacts}
Our work tackles the challenge of disentangling clothing and body shape in $3$D shape representation. The clothing reconstruction task remains challenging as evidenced by the visual quality of the published models and our models. Our results show that the task of body-clothing disentanglement brings benefit in the recognition task, a finding which opens new possibilities to multi-task learning across $2$D-$3$D modalities. Like most person ReID
methods, one potential negative impact of our approach is that it could be used for unethical surveillance and invasion of privacy.

\Paragraph{Acknowledgments} This research is based upon work supported by the Office of the Director of National Intelligence (ODNI), Intelligence Advanced Research Projects Activity (IARPA), via 2022-21102100004. The views and conclusions contained herein are those of
the authors and should not be interpreted as necessarily representing the official policies, either expressed or implied, of ODNI, IARPA, or the U.S. Government. The U.S. Government is authorized to reproduce and distribute reprints for governmental purposes notwithstanding any copyright annotation therein.